# Robotic Applications in Cardiac Surgery

**Alan P. Kypson & W. Randolph Chitwood Jr.**

East Carolina University, The Brody School of Medicine, Room 252, Division of Cardiothoracic Surgery,
600 Moye Boulevard, Greenville, N.C. 27858 USA.
E-mail: kypsona@mail.ecu.edu

*Abstract*: Traditionally, cardiac surgery has been performed through a median sternotomy, which allows the surgeon generous access to the heart and surrounding great vessels. As a paradigm shift in the size and location of incisions occurs in cardiac surgery, new methods have been developed to allow the surgeon the same amount of dexterity and accessibility to the heart in confined spaces and in a less invasive manner. Initially, long instruments without pivot points were used, however, more recent robotic telemanipulation systems have been applied that allow for improved dexterity, enabling the surgeon to perform cardiac surgery from a distance not previously possible. In this rapidly evolving field, we review the recent history and clinical results of using robotics in cardiac surgery.

## 1. Introduction

Since the advent of laparoscopic cholecystectomy, various surgical disciplines have successfully adopted endoscopic technology, leading to decreased morbidity and shorter recovery times. Most of these operations have involved the removal of an organ rather than the delicate construction of a coronary bypass anastomosis or valve repair. This is likely due to the fact that long, nonarticulated instruments with fixed pivot points that are the cornerstone of endoscopic surgery significantly limited dexterity. Furthermore, the lack of depth perception with traditional video systems has also made delicate reconstructive surgery difficult. Therefore, cardiac surgery has been slow to adopt such endoscopic techniques. With the development of robotic surgical systems surgeons have begun to explore the possibility of altering current approaches to cardiovascular surgery.

Cardiac surgery has usually been, and in most instances continues to be, performed via a median sternotomy. This incision allows direct and ample access to the heart and all of its surrounding structures. Thus, the surgeon is able to readily cannulate the structures necessary to initiate cardiopulmonary bypass (CPB). Furthermore, the exposure allows the surgeon to perform coronary artery bypass grafting (CABG), as well as intracardiac valve repair and replacement.

Recently, there has been a shift in the way cardiac surgery is performed. An extensive effort is underway to develop and perform less invasive cardiac surgery through incisions other than the classic median sternotomy. This changing paradigm raises a number of issues. No longer can the surgeon readily access the thoracic cavity to place cannulae for CPB or perform surgical maneuvers manually. In fact, new techniques for remote cannulation have been developed (Pompili *et al.*, 1996; Stevens *et al.*, 1996) Furthermore, with the changes in size and location of incisions, visualization systems and robotic telemanipulation systems have had to be developed to allow the surgeon to perform precise surgical maneuvers that mimic human dexterity. This article will review the history of robotic applications in cardiac surgery and highlight our early clinical experience.

## 2. History of Robotic Systems

Six degrees of freedom are required to allow free orientation in space. Thus, standard endoscopic instruments with only four degrees of freedom dramatically reduce dexterity. When working through a fixed entry point, such as a trocar, the operator must reverse hand motions (the fulcrum effect). At the same time, instrument shaft shear, or drag, induces the need for higher manipulation forces, leading to hand muscle fatigue. Also, human motor skills deteriorate with visual-motor incompatibility, which is commonly associated with endoscopic surgery. Computer-enhanced instrumentation systems have been developed to overcome these and other limitations. These systems provide both telemanipulation and micromanipulation of tissues in confined spaces. The surgeon operates from a console, immersed in a three-dimensional view of the operative field. Through a computer interface, the surgeon's motions are reproduced in scaled proportion through "microwrist" instruments that are mounted on robotic arms inserted through the chest wall. These instruments emulate human X-Y-Z axis wrist activity throughout seven full degrees of freedom.

During the early 1990s, extensive research conducted on robotic applications in surgery resulted in the emergence of two robotic systems. Based on technology developed by the Stanford Research Institute, Frederic Moll, MD, Robert Younge, and John Freund, MD developed the da Vinci System (Intuitive Surgical, Mountain View, CA) in 1995. The system stressed the telepresence concept, with the surgeon immersed in a full three-dimensional experience that created the sense that the operative site was directly in view. The hand motions exactly replicated the motions of open surgery through the use of "endowrists." As a result, the learning curve was minimized. The Computer Motion (Santa Barbara, CA) system, Zeus was developed as an integrated robotic surgical system. Yulun Wang, PhD, founded the company in 1989 and initially developed a voice-controlled robotic arm called the Automated Endoscopic System for Optimal Positioning (AESOP), which would be used as a laparoscopic camera holder. The Zeus system did not attempt to create an immersive intuitive interface; rather, the surgeon was fully aware of performing a robotic procedure. The fulcrum effect of laparoscopic surgery was retained, as well as the obvious monitor for viewing.

### 3. AESOP Robotic System

In 1994, AESOP became the first medical robot to receive Food and Drug Administration (FDA) approval in the United States. Two years later, voice control by the surgeon was added, allowing for exact, hands-free control of an endoscope during an operation. This robotic arm is capable of responding to over 20 simple voice commands. In 1996, while general surgeons were already using this device, Carpentier performed the first video-assisted mitral valve repair via a minithoracotomy using ventricular fibrillation (Carpentier *et al.*, 1996). Three months later, our group at East Carolina University (ECU) completed a mitral valve replacement using a microincision, videoscopic vision, percutaneous transthoracic aortic clamp, and retrograde cardioplegia (Chitwood *et al.*, 1997). In 1997, Mohr first used the AESOP voice-activated camera robot in minimally invasive videoscopic mitral valve surgery (Mohr *et al.*, 1998). Microincisions or port incisions (4 cm) were used and most of the operation was performed via secondary or assisted vision. In June 1998, our group performed the first completely video-directed mitral operation in the United States using AESOP and a Vista (Vista Cardiothoracic Systems, Inc., Westborough, MA) three-dimensional camera (Chitwood *et al.*, 1997). Continued experience, advances in three-dimensional videoscopy, and voice activated camera control with AESOP allowed the use of even smaller incisions with better visualization of both valvular and subvalvular components of the mitral valve. The voice-activated camera provides direct eye-brain-action translation for the surgeon without an intermediary assistant. Camera manipulation requirements are diminished, movements are more predictable, and lens cleaning is reduced, enabling the surgeon more operative flexibility and speed.

### 4. Zeus Robotic System

The Zeus system is composed of three interactive arms, mounted directly on the operating table compared with the da Vinci, which is positioned on the floor next to the patient (Figs. 1A and 1B). The surgeon's movements are digitalized and filtered by a signal processor before being relayed to the robotic arms. The surgeon is seated at a remote console, and handles provide a sensitive robotic interface. The system mechanically relays the surgeon's hand movements to a computer controller. The basic Zeus visualization system is two-dimensional; however, it can be used in combination with new and independently developed three-dimensional visualization systems. The surgeon remains seated with the endoscopic image displayed at eye level and close to the hands. Although the Zeus system lacks a fully articulated wrist and allows only four degrees of freedom, the instrument diameter is a small 3.9 mm compared with the 7 mm da Vinci arm. Recently, some instrument tips have been developed that provide five degrees of freedom through the use of a "microwrist." More than 20 different end-effectors exist within this system, including needle drivers, tissue graspers, and microscissors. These instruments are easily interchangeable during the surgical procedure. The arms on which these instruments are mounted are lightweight and flexible, allowing the surgical team to stand close to the patient.

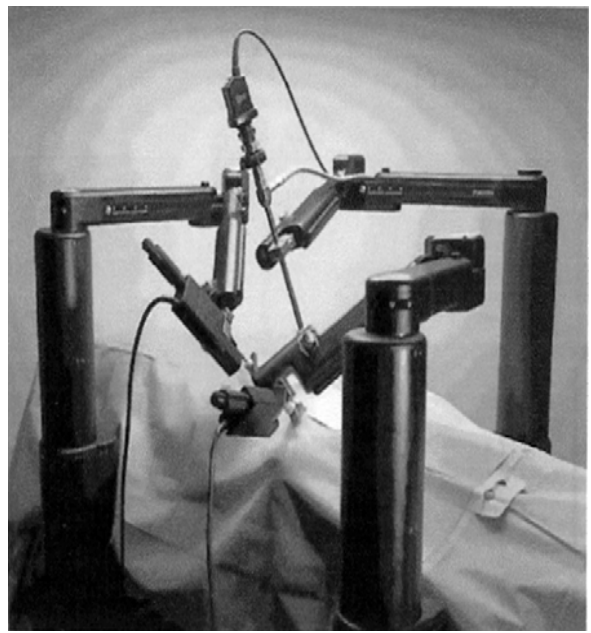

Fig. 1A. Zeus tableside robotic system.

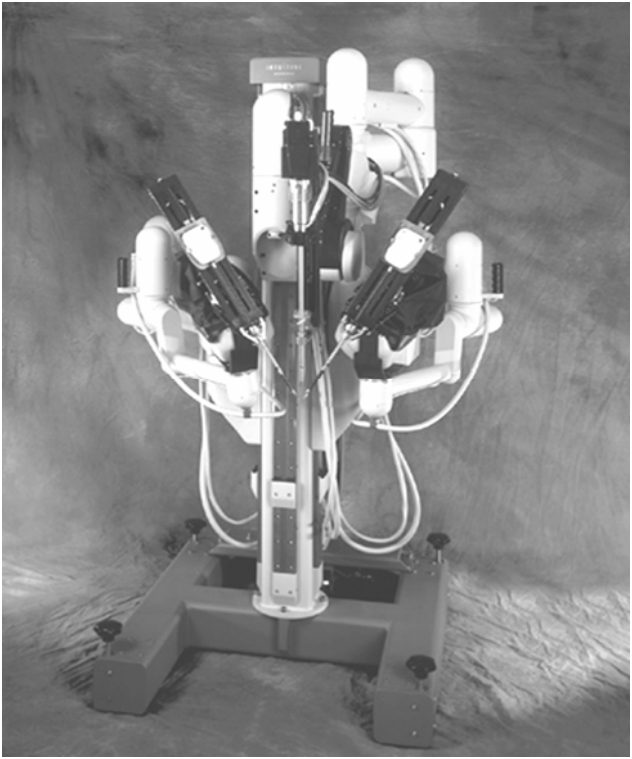

Fig. 1B. da Vinci tableside robotic system.

## 5. da Vinci Robotic System

Three components comprise the da Vinci system: a surgeon console, an instrument cart, and a visioning platform. The operative console is physically remote from the patient and allows the surgeon to sit comfortably, resting the arms ergonomically with his or her head positioned in a three-dimensional vision array. Through sensors, the surgeon's finger and wrist movements are digitally registered in computer memory banks; these actions are efficiently transferred to an instrument cart, which operates the synchronous end-effector instruments (Fig. 2). Wrist-like instrument articulation precisely emulates the surgeon's actions at the tissue level and enhances dexterity through combined tremor suppression and motion scaling. A clutching mechanism enables readjustment of hand positions to maintain an optimal ergonomic attitude with respect to the visual field.

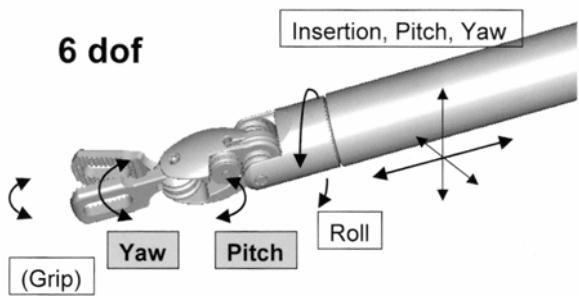

Fig. 2. End-effector arms of da Vinci system demonstrating multiple axis of rotation.

This clutch acts very much like a computer mouse, which can be reoriented by lifting and repositioning it to re-establish unrestrained freedom of computer activation. The three-dimensional digital visioning system enables natural depth perception with high-power magnification (10X). Both 0° and 30° endoscopes can be manipulated to look either "up" or "down" within the heart. Access to and visualization of the internal thoracic artery, coronary arteries, and mitral apparatus is excellent. The operator becomes ensconced in the three-dimensional operative topography and can perform extremely precise surgical manipulations devoid of traditional distractions.

Through 1 cm ports, instruments are positioned near the cardiac operative site in the thorax, and the camera is passed via a 4 cm working port used for suture and prosthesis passage (Fig. 3). Every analog finger movement, along with inherent human tremor at 8–10 Hz/second, is converted to binary digital data, which are smoothed and filtered to increase micro instrument precision. The wrist-like articulation and motion suppression allow both increased precision and dexterity with the surgeon becoming truly ambidextrous. Six degrees of freedom are offered by this combination of trocar-positioned arms (insertion, pitch, and yaw) and articulated instrument wrists (roll, grip, pitch, and yaw).

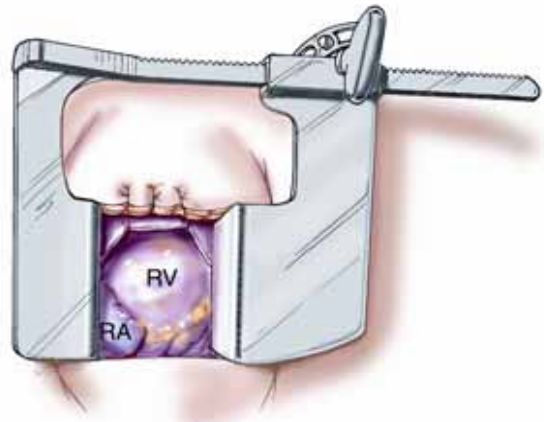

Fig. 3. 4-cm incision in the right 4th intercostal space.

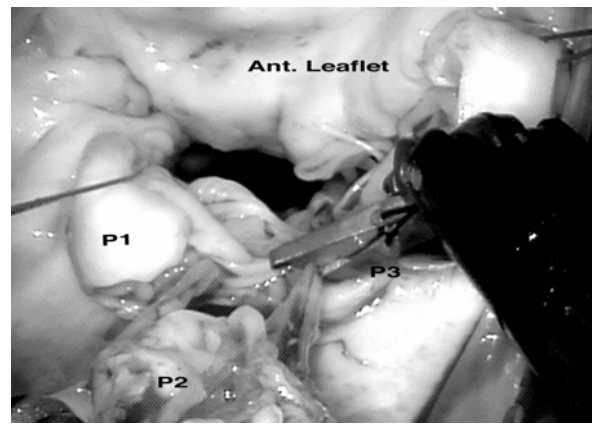

Fig. 4. View with the da Vinci system of the mitral valve.

Full X-Y-Z axis agility is affected by coordinating foot pedal clutching and hand motion sensors. Hand activity at the console is exactly reproduced at the surgical field while foot pedals control the camera position and focus. Coordination of these eye-hand-foot movements enables the surgeon to ratchet articulated wrists smoothly through every coordinate, allowing for complex instrument positions while providing maximum ergonomic comfort. Fig. 4 shows the surgeon's operative field during a da Vinci mitral repair.

## 6. Robotics in Cardiac Valve Surgery

In 2001, Felger reported on 72 robotically directed mitral valve surgeries performed at ECU with the AESOP system. These patients were compared to 55 manually directed patients (an assistant directed by the surgeon) undergoing mitral valve surgery (Felger *et al.*, 2001). Perioperative data analysis revealed that 75% of the robotically directed group underwent mitral valve repair, which included quadrangular resections, sliding plasties, chord replacement/repair, and Alfieri-plasties. Despite the complexity of these cases, robotically directed cross-clamp times were significantly less than the manually directed cohort ($90.0 \pm 4.6$ minutes versus $128.0 \pm 4.5$ minutes; $p<0.001$). Cardiopulmonary bypass times were less in the robotically directed group ($143.3 \pm 4.6$ minutes versus $172.8 \pm 5.7$ minutes; $p<0.001$). This time difference reflects the importance of robotic assistance, as less time was required for positioning the endoscope and cleaning the lens, enabling the surgeon to perform the operation more expeditiously. Comparative lengths of hospital stay were the same for the robotically directed and manually directed groups, but significantly less than a historical control of conventional sternotomy mitral surgery patients ($4.6 \pm 0.3$ days vs. $7.9 \pm 0.6$ days; $p<0.001$). There were two (3%) conversions to sternotomy secondary to bleeding in the robotically directed group. There were no permanent neurologic events in this group and no patients underwent prolonged ventilation (>48 hours). Thirty-day operative mortality was 2.3%, demonstrating both the efficacy and safety of this approach.

Other groups have reported similar success using AESOP with minimally invasive mitral valve surgery. Mishra and colleagues from the Escorts Heart Institute in India reported their experience in 221 patients, in which the AESOP system was used in 120 patients (Mishra *et al.*, 2002). Perfusion and aortic cross-clamp times were $126 \pm 41$ minutes and $58 \pm 16$ minutes, respectively. When compared to a conventional incision cohort, there was no significant difference. Most of these operations (81%) involved valve replacement accounting for the shorter perfusion times when compared with our series where only 25% of mitral valve operations were replacements. Excellent clinical results were achieved with no permanent neurologic deficits and an operative mortality of 0.45%. There were no reoperations in a follow-up period of 16 months, again demonstrating the safety and efficacy of robotically assisted minimally invasive mitral valve surgery.

In May 1998, using an early prototype of the da Vinci articulated "microwrist," Carpentier performed the first true robotic mitral valve repair (Carpentier *et al.*, 1998). In May 2000, our group performed the first complete repair of a mitral valve in North America using the da Vinci system (Chitwood *et al.*, 2000). With articulated wrist instruments, a trapezoidal resection of a large $P_2$ scallop was performed with the defect closed using multiple interrupted sutures, followed by implantation of a #28 Cosgrove annuloplasty band. The same year we completed an FDA safety and efficacy clinical trial in 10 patients. Quadrangular leaflet resections, leaflet sliding plasties, chord transfers, polytetrafluoroethylene chord replacements, reduction annuloplasties, and annuloplasty band insertions were all successfully performed. The mean total arrest time was 150 minutes, with 52 minutes used for leaflet repairs. Of the total arrest time, a mean of 42 minutes was needed to place an average of 7.5 annuloplasty band sutures. Total operating room time averaged 4.8 hours. There were no device-related complications, and only one re-exploration for bleeding from an atrial pacing wire. The average postoperative stay was 4 days (range 3-7 days). At three-month follow-up, echocardiography revealed nothing more remarkable than trace mitral regurgitation. All patients returned to normal activity by one month after surgery (Chitwood WR Jr & LW., 2001).

We recently published our results of the first 38 mitral repairs with the da Vinci system (Nifong *et al.*, 2003). For data analysis and comparison, patients were divided into two cohorts of 19 patients (early experience and late experience). Total robotic time represents the exact time of robot deployment after valve exposure and continued until the end of annuloplasty band placement. This time decreased significantly from $1.9 \pm 0.1$ hours in the first group to $1.5 \pm 0.1$ hours in the second group ($p=0.002$). Concurrently, leaflet repair time fell significantly from $1.0 \pm 0.1$ hours to $0.6 \pm 0.1$ hours, respectively ($p=0.004$). Also, total operating time decreased significantly from $5.1 \pm 0.1$ hours to $4.4 \pm 0.1$ hours in the second group of patients. Furthermore, both cross-clamp and bypass times decreased significantly with experience. Similar time trends were reported in a later publication that reviewed subsequent patients that underwent robotic mitral repair at our institution with the da Vinci system (Kypson *et al.*, 2003). The only time that did not change between the two groups was the annuloplasty band placement time. Most likely, this represents an inherent limitation in the current technology of tying suture knots. Even with extensive experience, the speed of suture placement is limited. Alternative methods of implanting the annuloplasty ring may decrease this time in the future. For the entire group of 38 patients, the mean length of stay was 3.8 days, with no difference between the two groups. For all patients in the study, 84% demonstrated a reduction of three or more grades in mitral regurgitation at follow-up. In the entire series there were no device-related complications or operative deaths. One valve was replaced at 19 days because of hemolysis secondary to a leak that was directed against a prosthetic chord.

Besides our own experience, results from a prospective multicenter phase II FDA trial were recently submitted (Chitwood Jr WR *et al, in press.*). In this trial involving 10 institutions, the da Vinci system was used to perform mitral valve repairs in 112 patients. Valve repairs included quadrangular resections, sliding-plasties, edge-to-edge approximations, and both chordal transfers and replacements. Leaflet repair times averaged 36.7 ± 0.2 minutes with annuloplasty times of 39.6 ± 0.1 minutes. Total robotic, aortic cross-clamp, and cardiopulmonary bypass times were 77.9 ± 0.3 minutes, 2.1 ± 0.1 hours, and 2.8 ± 0.1 hours, respectively. At one-month follow-up, transthoracic echocardiography revealed nine patients (8.0%) with greater than or equal to grade 2 mitral regurgitation and six (5.4%) of these had re-operations (five replacements, one repair). There were no deaths, strokes, or device-related complications. This study demonstrated that multiple surgical teams could perform robotic mitral valve surgery safely early in the development of this technique.

**7. Robotics in Coronary Artery Bypass Surgery**

In May 1998, Mohr and Falk harvested the left internal mammary artery (LIMA) with the da Vinci system and performed the first human coronary anastomosis through a small left anterior thoracotomy incision (Mohr *et al.*, 1999; Falk *et al.*, 2000). More recently, work with da Vinci has lead to FDA approval for internal mammary harvesting in the United States.

The first totally endoscopic coronary artery bypass (TECAB) was performed on an arrested heart at the Broussais Hospital in Paris using an early prototype of the da Vinci system (Loulmet *et al.*, 1999). The Leipzig group attempted a total closed chest approach for LIMA to left anterior descending coronary artery (LAD) grafting on the arrested heart in 27 patients with da Vinci and was successful in 22 patients (Falk *et al.*, 2000). Furthermore, surgeons in Europe improved the initial da Vinci coronary bypass graft method and eventually were able to complete off-pump bilateral grafts of the internal mammary artery to the anterior descending and right coronary arteries while working from one side of the chest (Aybek *et al.*, 2000; Kappert *et al.*, 2000).

Experience with endoscopic coronary artery bypass surgery has been limited to only a few centers and results are highly controlled. Because the success of coronary surgery depends on multiple, complex steps culminating in the creation of a vascular anastomosis, most clinical series have introduced robotically assisted coronary surgery in a stepwise fashion. Specifically, initial experience is limited to endoscopic LIMA harvesting, followed by a robotically assisted anastomosis through a median sternotomy. Subsequently, a total endoscopic procedure is performed on an arrested heart, and finally during a beating heart operation. Currently, the largest published series of TECAB comes from Europe. Wimmer-Greinecker's group in Frankfurt, Germany, reported on 45 patients who underwent TECAB on an arrested heart (Dogan *et al.*, 2002). Most of these (82%) were single-vessel bypass (either LIMA-LAD or right internal mammary artery to right coronary artery). The first 22 patients had angiograms prior to discharge, revealing a 100% patency rate. Mean operative time was 4.2 ± 0.9 hours for single-vessel TECAB and 6.3 ± 1.0 hours for double-vessel bypass procedures. The average cross-clamp time was 61 ± 16 minutes for single bypass and 99 ± 55 minutes for double bypass. The initial conversion rate of 22% decreased to 5% in the last twenty patients, reflecting an obvious learning curve.

In the United States, Damiano and his colleagues initiated a multicenter clinical trial on robotically assisted coronary surgery using the Zeus system (Damiano *et al.*, 2000). In this FDA safety and efficacy trial, 19 patients underwent a median sternotomy with cardioplegic arrest of the heart. All grafts were hand sewn in a traditional manner except the LIMA-LAD, which was robotically sewn. Seventeen patients had adequate intraoperative flow (mean 38.5 ± 5 mL/min) in the LIMA graft. Anastomotic time was 22.5 ± 1.2 minutes. One patient underwent re-exploration for mediastinal hemorrhage. At eight weeks follow-up, angiography showed that all grafts were patent. The average hospital stay was 4.1 ± 0.4 days. Boyd and associates from London Health Sciences Center in Ontario, Canada, have also been extensively involved in initial endoscopic coronary surgery trials with the Zeus system. In 2000, they published a series of six patients that were the first to undergo TECAB in North America on a beating heart using a specialized endoscopic stabilizer (Boyd *et al.*, 2000). Each of these patients had single vessel LAD disease and underwent LIMA-LAD grafting. Special 8-0 polytetrafluoroethylene suture 7 cm in length was used to minimize the time required for suture placement. Intracoronary shunts were used to provide needle depth landmark when performing endoscopic anastomosis with two-dimensional cameras. LIMA harvest time averaged 65.3 ± 17.6 minutes (range 50-91 minutes). The anastomotic time was 55.8 ± 13.5 minutes (range 40-74 minutes) and median operative time was 6 hours (range 4.5-7.5 hours). All patients had angiographically confirmed patent grafts before leaving the hospital. The average hospital length of stay was 4.0 ± 0.9 days.

**8. Conclusion**

The early clinical experience with computer-enhanced telemanipulation systems has defined many of the limitations of this approach despite rapid procedural success. Currently, the lack of force feedback is being addressed and a strain sensor is being incorporated into advanced robotic surgical tools and may soon allow more control of force applied at the robotic end-effector (Zenati, 2001). Furthermore, conventional suture and knot tying add significant time to each procedure. Technologic advancements, such as the use of nitinol U-clips (Coalescent Surgical Inc., Sunnyvale, CA) (Fig. 5) instead of sutures requiring manual knot tying, should decrease operative times significantly.

Besides advances in surgical technology, the potential use of image-guided surgical technologies will provide real-time data acquisition of physiologic characteristics, allowing one to better assess the delivery of remote percutaneous therapy. Imaging techniques may include three-dimensional modeling and reconstruction from computerized tomography, magnetic resonance imaging, or ultrasound.

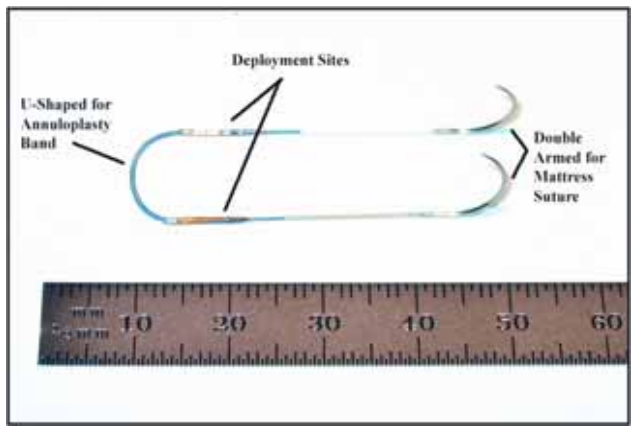

Fig. 5. Nitinol U-clips.

Recently published research conducted at the National Research Institute in Computer Science and Control of France, in collaboration with Professor Carpentier, focuses on simulating and planning robotic procedures (Blondel *et al., 2002*). Computer modeling of organs such as the heart is generated from combined information from different imaging modalities. A surgeon may visualize and manipulate simulated objects interactively and once optimal access port placements are determined, the positions of the simulated tools can be recorded and marked directly on the patient to specify positions for port incisions. This technology should enable and facilitate totally endoscopic robotic cardiac operations and benefit the patient through decreased risks and fewer adverse operative outcomes.